\documentclass[a4paper,10pt,conference]{ieeeconf}

\usepackage{FG2027}
\FGfinaltrue

\IEEEoverridecommandlockouts
\def\FGPaperID{****}

\usepackage{amsmath}
\usepackage{amssymb}
\usepackage{graphicx}
\usepackage{booktabs}
\usepackage{array}
\usepackage{tabularx}
\usepackage{xcolor}
\usepackage{cite}
\usepackage{hyperref}
\usepackage{placeins}
\usepackage{float}

\hypersetup{hidelinks}
\graphicspath{{images/}}

\title{\LARGE \bf
MEGMAG: Micro-Expression Generation via MAGnification
}
\author{\parbox{16cm}{\centering
    {\large
    \href{https://orcid.org/0009-0008-1846-7414}{Sliman Jammal} and
    \href{https://orcid.org/0000-0002-3963-4508}{Andrei Sharf}}\\
    {\normalsize Ben-Gurion University of the Negev, Israel}\\
    {\normalsize\tt sliman@post.bgu.ac.il, asharf@bgu.ac.il}}
}

\begin{document}
\thispagestyle{empty}
\pagestyle{empty}
\maketitle

\begin{abstract}
Facial micro-expressions are subtle, short-lived facial movements that reveal genuine human emotions. Despite the large advances in facial expression synthesis and animation, realistic micro-expressions remain challenging due to scarce annotated data and the extremely subtle nature of their motion. Existing dedicated micro-expression methods therefore often struggle with generation quality, robustness, and generalization.

We present~\textbf{MEGMAG}, a transferable framework for micro-expression generation via magnification that enables pretrained facial animation models to generate realistic micro-expressions without modifying or fine-tuning their backbones. The key idea is to transform subtle micro-expression dynamics into the domain of regular facial expressions, where pretrained animation models can effectively capture and transfer facial motion.
A learnable \textbf{Magnification Module} amplifies subtle micro-expression dynamics, enabling a frozen facial animation backbone to transfer the amplified motion to a source identity without altering its original architecture or parameters. 
A~\textbf{DeMagnification Module} then attenuates the generated motion to recover realistic micro-expression characteristics. This design allows MEGMAG to leverage diverse off-the-shelf facial animation models while remaining architecture-agnostic and fully transferable across backbones.

We evaluate MEGMAG with three pretrained, frozen facial animation backbones—FOMM, MetaPortrait, and EmoPortraits—on CAS(ME)$^3$ and SAMMLV datasets. Recognition is assessed using BoostingVRME-based models trained exclusively on real micro-expression data, while spotting is evaluated on complete SAMMLV sequences. 
On CAS(ME)$^3$, MEGMAG consistently improves UAR, UF1, and Accuracy across all three backbones. On SAMMLV, it improves full-video spotting F1 and STRS for every backbone and nearly all recognition measures. Our EmoPortraits-based model achieves the strongest recognition performance on CAS(ME)$^3$ and outperforms existing dedicated micro-expression generation methods. Overall, these results demonstrate that learned micro-expression magnification provides an effective and transferable bridge between conventional facial animation and micro-expression generation.

\end{abstract}

\begin{figure}[!t]
    \centering
    \includegraphics[width=1\linewidth]{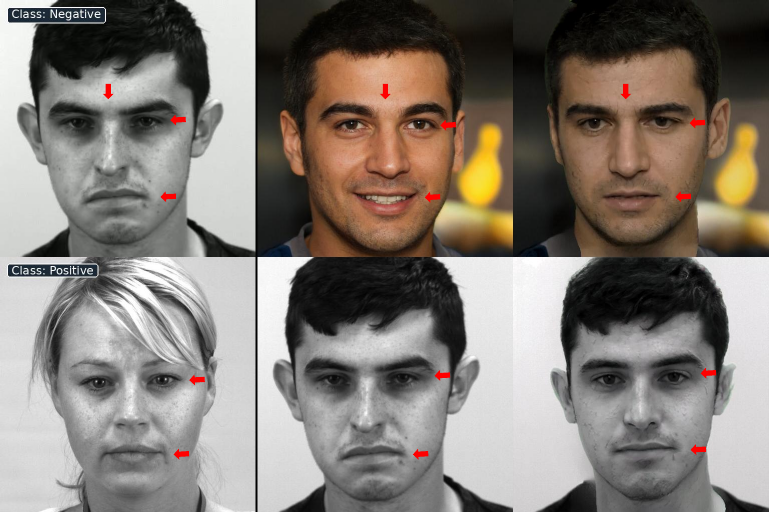}
    \caption{MEGMAG micro-expression generation results. Each row shows, from left to right, the driving frame, the source identity image, and the generated micro-expression transferred to the source identity Red arrows highlight subtle facial movements.}
    \label{fig:teaser}
\end{figure}

\section{Introduction}
\label{sec:introduction}

Facial micro-expressions are involuntary, subtle, and short-lived facial movements that reveal concealed emotional states, typically lasting less than 0.5 seconds with extremely low motion intensity~\cite{Ben2022MicroExpressionSurvey}. Despite their subtlety, they provide valuable cues for genuine emotion and have applications in affective computing, emotion recognition, clinical psychology, and human--computer interaction. However, their extremely low motion magnitude can fall below the sensitivity of feature spaces learned from macro-expression dynamics, causing subtle facial displacements to be treated as noise by existing generative frameworks. Consequently, generating realistic and temporally coherent micro-expressions remains challenging.

Another major challenge is the severe scarcity of annotated micro-expression data. Building such datasets requires controlled elicitation protocols, high-frame-rate recording, and labor-intensive annotation by certified FACS coders~\cite{Ekman2002FACS}. Consequently, existing datasets~\cite{davison2018samm,yap2020sammlv,Li2023CASME3} remain limited in scale, hindering the scalability and generalization of dedicated end-to-end approaches. As a result, existing micro-expression models often exhibit limited synthesis quality and poor cross-subject and cross-expression generalization.

In contrast, macro-expressions last longer and involve substantially larger facial deformations~\cite{Ben2022MicroExpressionSurvey}. Accordingly, recent large-scale facial animation models are predominantly trained on regular expressions, enabling impressive photorealistic expression synthesis and motion-driven portrait animation~\cite{Drobyshev2024EMOPortraits,rochow2024fsrt,hong2022depth}, but creating a representational bias toward large-scale facial dynamics that limits their sensitivity to the subtle, brief, and spatially localized motion of micro-expressions. We bridge this macro-to-micro gap with a magnification-driven, backbone-agnostic framework that reformulates micro-expression synthesis as a motion-domain transfer problem. Rather than modifying or fine-tuning pretrained animation models—which is challenging under severe data scarcity—we magnify subtle micro-expression dynamics to enable pretrained models to reliably capture and transfer facial motion.

We introduce a learned \textbf{Magnification Module} based on FlowMag~\cite{Pan2023SelfSupervised}, repurposing motion magnification for micro-expression generation. Unlike prior work that focuses on amplifying subtle motion for visualization, MEGMAG uses magnification to bridge the motion-domain gap between micro-expressions and macro-trained facial animation backbones. The module amplifies subtle micro-expression dynamics, allowing frozen pretrained backbones to transfer them to a source identity without architectural changes or fine-tuning. A complementary \textbf{DeMagnification Module} then attenuates the transferred motion to realistic micro-expression intensity. This magnify--generate--demagnify pipeline turns motion magnification into a practical mechanism for backbone-agnostic micro-expression synthesis while preserving identity and subtle facial dynamics (Figure~\ref{fig:teaser}).

Our contributions are:
\begin{itemize}
    \item \textbf{Backbone-Agnostic Generation.} A framework for micro-expression generation using pretrained facial animation backbones without modifying or fine-tuning them.

    \item \textbf{Learned Magnification.} A module that amplifies subtle micro-expression dynamics for effective motion transfer.

    \item \textbf{Learned DeMagnification.} A module that restores realistic motion intensity while preserving fine-grained dynamics.
\end{itemize}

\begin{figure*}[t]
    \centering
    \includegraphics[width=\textwidth]{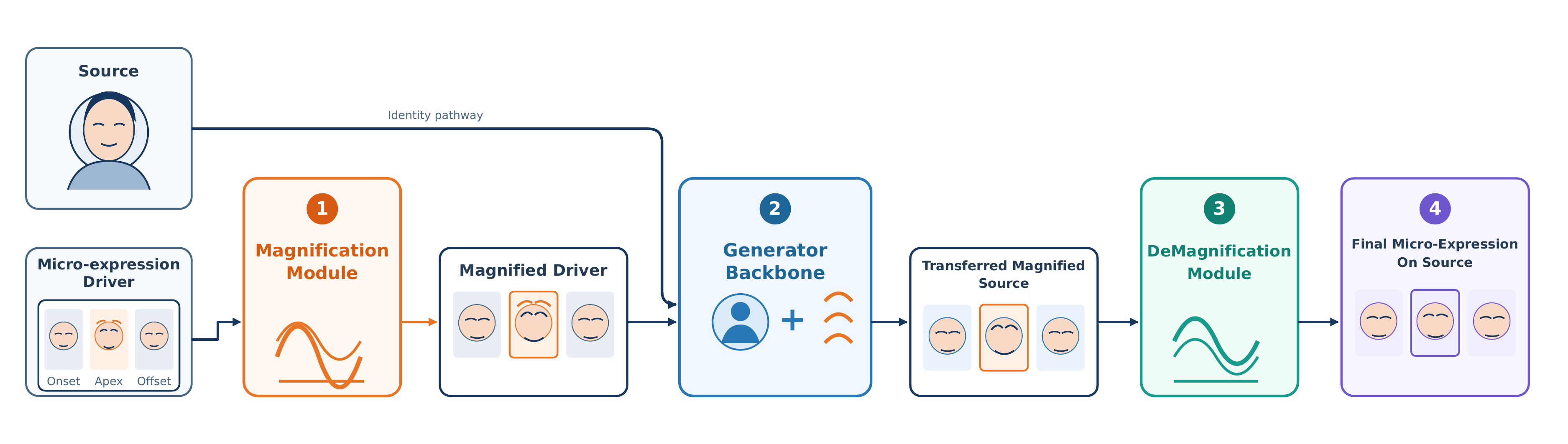}
    \caption{Pipeline illustration. Left to right: the driver micro expression video is processed by the Magnification Module (1) and then provided together with the source identity image to a frozen facial generator backbone (2). The generated magnified sequence is processed by the DeMagnification Module (3), resulting in a transferred micro expression on the source identity (4).}
    \label{fig:pipeline}
\end{figure*}

\section{Related Work}
\label{sec:related}

We review recent advances in facial expression synthesis, micro-expression generation, and motion magnification.

\textbf{Facial Expression Synthesis.}
Facial expression generation has advanced rapidly, driven by large-scale datasets and increasingly powerful generative architectures. Image animation methods synthesize videos that transfer motion from a driving sequence to a source image while preserving identity.

Early approaches relied on explicit facial priors, such as landmarks and facial boundaries~\cite{Wiles2018X2Face,Siarohin2019MonkeyNet}. Subsequent self-supervised methods learned motion representations directly from data. Notable examples include FOMM~\cite{Siarohin2019FOMM}, which introduced unsupervised keypoint-based motion transfer; MRAA~\cite{Siarohin2021MRAA}, which extended motion modeling to articulated objects; and thin-plate spline approaches~\cite{Zhao_2022_CVPR}, which improved the flexibility of spatial deformation.

Recent neural portrait animation models~\cite{Wang2021FaceVid2Vid,Drobyshev2022MegaPortraits,Drobyshev2024EMOPortraits,rochow2024fsrt} achieve photorealistic synthesis with effective motion transfer and identity preservation. Models such as EmoPortraits~\cite{Drobyshev2024EMOPortraits} and FSRT~\cite{rochow2024fsrt} leverage large video corpora to learn expressive motion representations and generate realistic facial animations.

Despite these advances, such models are primarily developed for regular facial expressions, characterized by relatively large and sustained facial deformations. Their learned representations may therefore be less sensitive to the subtle, brief, and spatially localized movements of micro-expressions, limiting their ability to capture and transfer these low-amplitude dynamics.

\textbf{Micro-Expression Analysis and Generation.}
Micro-expression analysis primarily involves two tasks: micro-expression recognition (MER), which classifies segmented micro-expression clips, and spotting, which localizes micro-expressions in long videos. Unlike macro-expressions, micro-expressions are brief, involuntary, and low-intensity facial movements, making them challenging to detect and characterize reliably~\cite{Ben2022MicroExpressionSurvey}. Although public datasets have advanced research in this area~\cite{davison2018samm,li2013smic,yap2020sammlv,Li2023CASME3}, their limited scale and severe class imbalance remain major challenges.

Early MER methods relied on handcrafted motion descriptors, including LBP-TOP~\cite{Zhao2007LBPTOP}, optical-flow representations~\cite{Liu2016MDMO}, and histogram-based motion features~\cite{Davison2015MicroMovement}. Recent deep learning approaches learn spatiotemporal representations using optical flow, multi-stream architectures, and self-supervised motion learning~\cite{Gan2019OFFApexNet,liong2019ststnet,Guo2023GLEFFN,Fan2023SelfME,Zhang2025SODA4MER}. Micro-expression spotting has similarly progressed from handcrafted signal-processing pipelines to CNN-based and spatiotemporal methods for localizing subtle facial events in long videos~\cite{Zhang2018SMEConvNet,Zhang2020SpatioTemporalFusion,Li2019SpottingLongVideos,Guo2025BoostingVRME}. Recent approaches also investigate unified frameworks for joint spotting and recognition~\cite{Liong2023MEAN,Liong2024SFAMNet,Guo2025METSTPlus}.

Despite this progress, the scarcity of annotated micro-expression data remains a fundamental limitation. Dataset construction requires controlled elicitation protocols, high-frame-rate recording, and labor-intensive annotation by certified Facial Action Coding System (FACS)~\cite{Ekman2002FACS} coders. Consequently, limited training data constrain the development of robust generative models and their generalization to unseen subjects and expressions.

Research on micro-expression generation has explored adapting existing facial animation frameworks. Zhang et al.~\cite{zhang2021fpbfomm,zhang2024tipfacial} introduced facial-prior modules for FOMM~\cite{Siarohin2019FOMM} and MRAA~\cite{Siarohin2021MRAA}, while the MEGC2022 challenge~\cite{Li2022MEGC2022} featured approaches based on face-parsing constraints~\cite{Yu_2022_ACMMM} and dual-stream motion extraction~\cite{Fan_2022_ACMMM}. Although these methods improve the representation of subtle facial motion, they generally modify the underlying animation architecture and require fine-tuning on scarce micro-expression data, limiting their transferability across different backbones.

\textbf{Motion Magnification.}
Motion magnification amplifies subtle temporal variations to reveal otherwise imperceptible motion. Classical methods, such as Eulerian Video Magnification (EVM)~\cite{Wu2012EVM}, amplify temporally bandpass-filtered signals at fixed pixel locations. Wadhwa et al.~\cite{Wadhwa2013PhaseBased} instead amplify phase variations in a multiscale complex steerable pyramid representation. Subsequent work improves efficiency using Riesz pyramids~\cite{Wadhwa2014RieszPyramids} and separates foreground and background motion. Other approaches selectively amplify small motions using second-order temporal filters~\cite{Pan2023SelfSupervised}.
Despite these advances, Eulerian methods introduce noise and artifacts at high magnification levels. Moreover, they are not designed to preserve the sparse, localized, and characteristic dynamics of facial micro-expressions for generation.

Recent learning-based methods improve motion magnification by learning to amplify subtle dynamics while preserving image structure~\cite{Oh2018DeepMag,Pan2023SelfSupervised}. DeepMag~\cite{Oh2018DeepMag} learns magnification from synthetic training data, whereas FlowMag~\cite{Pan2023SelfSupervised} adopts a self-supervised approach using unlabeled videos. Other methods explore lightweight architectures and attention-based mechanisms, including micro-expression magnification~\cite{Kalluri2023FLAVR,Singh2023Lightweight}.

However, magnifying subtle general motions for visualization does not directly address the challenges of micro-expression generation. Micro-expressions involve sparse, localized, and brief facial movements that must be transferred to a different identity while preserving facial structure and motion characteristics. Excessive amplification can distort these dynamics, while insufficient amplification may leave them undetectable to facial animation models trained predominantly on macro-expressions. 

MEGMAG addresses this gap by magnifying micro-expression dynamics before motion transfer and applying demagnification afterward to restore realistic motion intensity.

\section{Method}
\label{sec:method}

Our framework enables micro-expression generation using a frozen pretrained facial animation backbone, preceded by a Magnification Module and followed by a DeMagnification Module. Given a source identity image and a driving video containing a micro-expression, the Magnification Module amplifies subtle facial dynamics to facilitate motion transfer by the pretrained backbone. The DeMagnification Module then attenuates the generated motion to restore realistic micro-expression intensity while preserving its fine-grained dynamics. The resulting \emph{magnify--generate--demagnify} pipeline enables micro-expression synthesis without modifying or fine-tuning the backbone (see Figure~\ref{fig:pipeline} for pipeline illustration).

\textbf{Magnification Module.}
The module learns to amplify subtle micro-expression dynamics toward the intensity range of regular facial expressions (Figure~\ref{fig:pipeline}, stage 1). Following Pan et al.~\cite{Pan2023SelfSupervised}, we adopt a self-supervised Lagrangian formulation that uses a frozen optical-flow estimator to provide motion supervision, avoiding the need for ground-truth magnified frames.

Given a reference frame $I_0$ and a subsequent frame $I_t$, the generator $G$ predicts a magnified frame
\[
\tilde{I}_t = G(I_0, I_t, \alpha),
\]
where $\alpha$ is the desired magnification factor. Let $\mathcal{F}(\cdot,\cdot)$ denote the optical flow estimated by the frozen network. The magnification loss enforces consistency between the generated motion and the scaled original motion:
\begin{equation}
\mathcal{L}_{\text{mag}}
=
\left\lVert
\alpha \mathcal{F}(I_0, I_t)
-
\mathcal{F}(I_0, \tilde{I}_t)
\right\rVert_1 .
\label{eq:mag_loss}
\end{equation}

Since optical flow alone does not fully constrain appearance, we introduce a color consistency loss. It compares the two frames after warping them into the reference coordinate system using their respective flow fields:
\begin{equation}
\mathcal{L}_{\text{color}}
=
\left\lVert
\mathrm{warp}\!\left(I_t, \mathcal{F}(I_0, I_t)\right)
-
\mathrm{warp}\!\left(\tilde{I}_t, \mathcal{F}(I_0, \tilde{I}_t)\right)
\right\rVert_1 .
\label{eq:color_loss}
\end{equation}

The combined training objective is
\begin{equation}
\mathcal{L}
=
w_{\text{mag}} \mathcal{L}_{\text{mag}}
+
w_{\text{color}} \mathcal{L}_{\text{color}},
\label{eq:total_loss}
\end{equation}
with $w_{\text{mag}}=1.0$ and $w_{\text{color}}=10.0$.

\begin{figure}[t]
    \centering
    \includegraphics[width=1\linewidth]{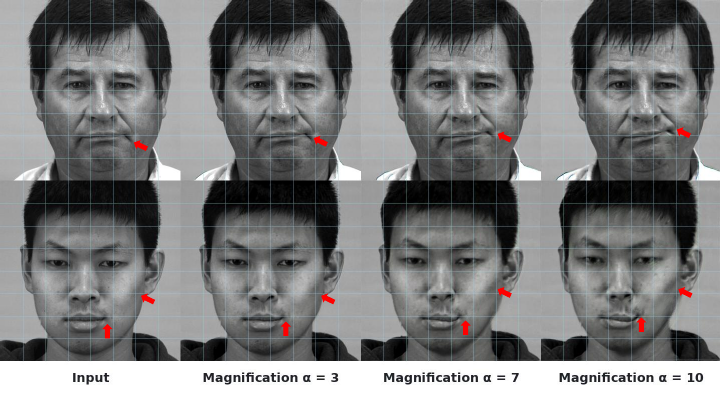}
    \caption{Demonstration of our Magnification Module on two micro-expression driver clips at magnification factors $\alpha \in \{3,7,10\}$. Increasing $\alpha$ progressively amplifies subtle facial displacements. Red arrows highlight magnified facial regions, and the overlaid grid provides a reference for assessing motion amplitude.}
    \label{fig:magplus_alpha_samm}
\end{figure}

The magnification generator follows a U-Net architecture~\cite{Ronneberger2015UNet} with five encoder--decoder stages, a base width of 64 channels that doubles at each downsampling stage, and skip connections between corresponding stages. The magnification factor $\alpha$ is encoded using a 32-dimensional sinusoidal positional embedding, spatially tiled and concatenated with the two input frames to form a 38-channel input tensor. Optical flow is estimated using a frozen RAFT model~\cite{Teed2020RAFT} with five refinement iterations. Figure~\ref{fig:magplus_alpha_samm} illustrates the effect of $\alpha_{\mathrm{mag}}$ on two SAMMLV apex frames.

\textbf{Facial Expression Transfer Backbone.}
The magnified driving frames and source identity image are fed into a frozen facial animation transfer backbone (Figure~\ref{fig:pipeline}, stages 1--2). Magnification brings subtle micro-expression dynamics into the motion range of regular facial expressions, enabling the backbone to capture and transfer them to the source identity.

Notably, the backbone transfers the driver's amplified facial motion to the source identity while preserving its appearance, without architectural modifications, expression-class-specific retraining, or fine-tuning.

\textbf{DeMagnification Module.}
The facial animation backbone transfers the magnified expression dynamics to the source identity, producing exaggerated motion dynamics that must be attenuated to the subtle intensity of natural micro-expressions. The DeMagnification Module (Figure~\ref{fig:pipeline} stage 3) learns this attenuation through the same self-supervised motion objective used for magnification, but with factors $\alpha<1$.

The DeMagnification Module uses the same U-Net architecture, optical-flow objectives (Eqs.~\eqref{eq:mag_loss}--\eqref{eq:total_loss}), and frozen RAFT estimator as the Magnification Module. The key difference is the sampling range of the attenuation factor. Whereas the Magnification Module samples $\alpha$ log-uniformly from $[1,16]$, the DeMagnification Module samples from the reciprocal range $[\tfrac{1}{16},1]$:
\begin{equation}
\log_2(\alpha)
\sim
\mathcal{U}\!\left(
\log_2\!\left(\tfrac{1}{16}\right),
\log_2(1)
\right)
=
\mathcal{U}(-4,0).
\label{eq:DeMagPlus_alpha}
\end{equation}

This objective encourages the output motion to have an amplitude scaled by $\alpha$ relative to the input motion. Applied after facial animation, the learned attenuation reduces exaggerated movements to realistic micro-expression intensity while retaining fine-grained dynamics and the source identity.

\section{Training}
\label{sec:training}

Our method trains only the Magnification and DeMagnification Modules, leaving the pretrained facial animation backbone frozen. This is substantially simpler than training a dedicated micro-expression generator: the modules are first trained on large-scale general video data and then fine-tuned on micro-expressions  to capture the subtle facial dynamics characteristic of this domain.

\textbf{Pretraining.}
We pretrain the FlowMag generator~\cite{Pan2023SelfSupervised} on approximately 145K frame pairs curated from publicly available video datasets, including YouTube-VOS~\cite{Xu2018YouTubeVOS}, DAVIS~\cite{PontTuset2017DAVIS}, Vimeo-90K~\cite{Xue2019Vimeo90K}, TAO~\cite{Dave2020TAO}, and UVO~\cite{Wang2021UVO}. Frame pairs with excessive or negligible motion are filtered using RAFT~\cite{Teed2020RAFT} flow statistics. Pretraining uses Adam~\cite{Kingma2015Adam}, a learning rate of $3\times10^{-4}$, and a resolution of $512\times512$.

The Magnification Module is pretrained with amplification factors sampled from $\alpha\in[1,16]$. The DeMagnification Module is initialized from the pretrained magnification weights and trained separately with attenuation factors sampled from $\alpha\in[\tfrac{1}{16},1]$.

\textbf{Fine-tuning on Micro-Expression Data.}
For each target dataset, we construct fine-tuning pairs from annotated micro-expression clips, sampling frames within their onset and offset boundaries. We use subject-independent splits to ensure that training and evaluation subjects do not overlap, thereby assessing generalization to unseen identities.

The Magnification Module is initialized from its forward pretrained checkpoint and fine-tuned on frame pairs from the training subjects using a learning rate of $10^{-4}$ and a batch size of 4--8, depending on the dataset. We retain the self-supervised losses and $\alpha$-conditioning strategy and update only the generator parameters. The DeMagnification Module is fine-tuned independently using its inverse pretrained checkpoint, the same subject split, and attenuation factors sampled from $[\tfrac{1}{16},1]$.

Both modules are adapted independently for each dataset and facial animation backbone. We select amplification and attenuation factors through validation sweeps and fix the selected values before testing; thus, optimal factors may vary across backbones. RAFT and all facial animation backbones remain frozen throughout training and evaluation.

\section{Experiments}
\label{sec:experiments}

\begin{figure*}[h!]
    \centering
    \includegraphics[width=0.64\linewidth]{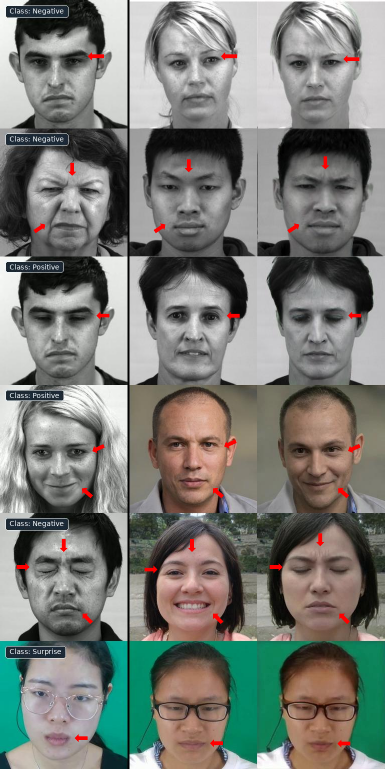}
    \caption{Qualitative results of micro-expression generation. From left to right: driving frame, source identity, and generated micro-expression transferred to source identity. Red arrows highlight regions containing subtle facial movements.}
    \label{fig:samm_qualitative}
\end{figure*}

\textbf{Experimental Setup.}
We evaluate MEGMAG on two public micro-expression datasets: SAMMLV~\cite{davison2018samm,yap2020sammlv} and CAS(ME)$^3$~\cite{Li2023CASME3}, which provide facial micro-expression videos with temporal and expression annotations. We consider three expression categories: Positive, Negative, and Surprise, and use subject-independent partitions to prevent identity overlap across training, validation, and test sets.
Figures~\ref{fig:teaser} and~\ref{fig:samm_qualitative} presents qualitative examples of the transferred micro expressions.

We evaluate three frozen facial animation backbones: FOMM~\cite{Siarohin2019FOMM}, MetaPortrait~\cite{Zhang2023MetaPortrait}, and EmoPortraits~\cite{Drobyshev2024EMOPortraits}. Each backbone is tested in two settings: the original backbone alone and the complete MEGMAG pipeline, which applies magnification before animation and demagnification afterward. The backbones remain frozen throughout and are never fine-tuned on micro-expression data.

The Magnification and DeMagnification Modules are fine-tuned independently for each dataset and backbone. Their amplification and attenuation factors are selected through validation sweeps and fixed before final testing.

On CAS(ME)$^3$, configuration selection uses a separate validation partition. Because official pretrained BoostingVRME weights are unavailable for this dataset, evaluation is limited to recognition using models trained exclusively on CAS(ME)$^3$. All methods use the same test split, and predictions are obtained by averaging logits from three models.

On SAMMLV, we process complete videos at the native frame rate, preserving all frames throughout magnification, animation, demagnification, and evaluation. The official pretrained BoostingVRME leave-one-subject-out (LOSO) models are used for recognition and spotting without retraining. Subjects used for configuration selection are excluded from the reported test results.

\textbf{Evaluation Protocol and Metrics}
Since no standardized evaluation protocol exists for micro-expression generation, we evaluate generated expressions indirectly through recognition and spotting~\cite{Liong2023MEAN,Fan2025MEGC2025}, using BoostingVRME~\cite{Guo2025BoostingVRME}, the top-ranked method in the MEGC2025 Spot-Then-Recognize challenge~\cite{Fan2025MEGC2025}. Recognition assesses whether generated clips preserve the intended expression category, while spotting measures their detectability in long, untrimmed videos. As official pretrained weights are available only for SAMMLV, we evaluate recognition on segmented CAS(ME)$^3$ clips and both recognition and spotting on complete SAMMLV videos at 200 fps.

\begin{table}[t]
\centering
\setlength{\tabcolsep}{2pt}
\resizebox{\columnwidth}{!}{%
\begin{tabular}{
l
@{\hspace{8pt}}c
@{\hspace{8pt}}c
@{\hspace{8pt}}c
@{\hspace{12pt}}c
@{\hspace{8pt}}c
@{\hspace{8pt}}c
}
\toprule

& \multicolumn{3}{c}{\textbf{CAS(ME)\textsuperscript{3}}}
& \multicolumn{3}{c}{\textbf{SAMMLV}} \\
\cmidrule(lr){2-4}\cmidrule(lr){5-7}
\multicolumn{1}{c}{\textbf{Method}}
& \textbf{UAR$\uparrow$}
& \textbf{UF1$\uparrow$}
& \textbf{Acc.$\uparrow$}
& \textbf{UAR$\uparrow$}
& \textbf{UF1$\uparrow$}
& \textbf{Acc.$\uparrow$} \\
\midrule

FOMM
& 0.3562 & 0.3460 & 0.4573
& 0.1429 & 0.1429 & 0.2727 \\

MEGMAG+FOMM
& \textbf{0.3856}
& \textbf{0.3737}
& \textbf{0.4874}
& \textbf{0.2593}
& \textbf{0.2333}
& \textbf{0.4375} \\

\midrule

MetaPortrait
& 0.3197 & 0.2929 & 0.4121
& 0.2222 & 0.1905 & \textbf{0.4000} \\

MEGMAG+MetaPortrait
& \textbf{0.3878}
& \textbf{0.3292}
& \textbf{0.4623}
& \textbf{0.3889}
& \textbf{0.4667}
& \textbf{0.4000} \\

\midrule

EmoPortraits
& 0.3962 & 0.3645 & 0.4523
& 0.2727 & 0.2609 & \textbf{0.6429} \\

MEGMAG+EmoPortraits
& \textbf{0.4344}
& \textbf{0.4125}
& \textbf{0.5276}
& \textbf{0.3077}
& \textbf{0.2857}
& 0.6316 \\

\bottomrule
\end{tabular}%
}
\caption{Comparison of micro-expression recognition performance on CAS(ME)\textsuperscript{3}~\cite{Li2023CASME3} and SAMMLV~\cite{yap2020sammlv}, using BoostingVRME~\cite{Guo2025BoostingVRME} with three frozen generator backbones, with and without MEGMAG. Results cover the Positive, Negative, and Surprise classes.}
\label{tab:recognition_casme3_sammlv}
\end{table}

\textbf{Micro-expressions Recognition}
We measure Unweighted Average Recall (UAR), Unweighted F1 (UF1), and Accuracy. UAR assigns equal weight to each expression class, making it suitable for imbalanced datasets. UF1 averages class-wise F1 scores, while Accuracy measures the proportion of correctly classified samples.

On CAS(ME)$^3$, all methods are evaluated on the same reviewed test split using three BoostingVRME-based models trained exclusively on real CAS(ME)$^3$ clips, with random seeds $13$, $42$, and $73$. We average their logits to obtain the final predictions.

On SAMMLV, we use the official pretrained BoostingVRME leave-one-subject-out (LOSO) models. Recognition is evaluated on correctly matched spotted events and reported alongside spotting performance.

Table~\ref{tab:recognition_casme3_sammlv} compares recognition performance across three frozen facial animation backbones, with and without the MEGMAG magnify--generate--demagnify pipeline.

On CAS(ME)$^3$, MEGMAG consistently improves all three recognition metrics across all backbones, with EmoPortraits achieving the strongest overall performance.

On SAMMLV, MEGMAG improves all metrics with FOMM. With MetaPortrait, it improves UAR and UF1 while maintaining Accuracy. With EmoPortraits, it improves UAR and UF1, while Accuracy decreases slightly from $0.6429$ to $0.6316$.

Overall, MEGMAG improves or preserves 17 of the 18 recognition measures across both datasets, demonstrating consistent benefits across different frozen generation backbones.

\begin{table}[t]
\centering
{\scriptsize
\setlength{\tabcolsep}{2pt}
\renewcommand{\arraystretch}{1.05}
\begin{tabularx}{\columnwidth}{
@{}
>{\raggedright\arraybackslash}X
>{\centering\arraybackslash}X
>{\centering\arraybackslash}X
>{\centering\arraybackslash}X
@{}
}
\toprule
\textbf{Method}
& \textbf{STRS$\uparrow$}
& \textbf{Spot. F1$\uparrow$}
& \textbf{Rec. F1$\uparrow$} \\
\midrule

FOMM
& 0.0104
& 0.0726
& 0.1429 \\

MEGMAG+FOMM
& \textbf{0.0267}
& \textbf{0.1146}
& \textbf{0.2333} \\

\midrule

MetaPortrait
& 0.0053
& 0.0279
& 0.1905 \\

MEGMAG+MetaPortrait
& \textbf{0.0233}
& \textbf{0.0474}
& \textbf{0.4912} \\

\midrule

EmoPortraits
& 0.0219
& 0.0838
& 0.2609 \\

MEGMAG+EmoPortraits
& \textbf{0.0384}
& \textbf{0.1345}
& \textbf{0.2857} \\

\bottomrule
\end{tabularx}
}
\caption{Comparison of full-video micro-expression spotting on SAMMLV~\cite{davison2018samm,yap2020sammlv} at the native 200 fps, using the official BoostingVRME~\cite{Guo2025BoostingVRME} LOSO models, with and without MEGMAG across three frozen generator backbones.}
\label{tab:sammlv_spotting}
\end{table}

\textbf{Micro-expressions Spotting}
We evaluate spotting on complete SAMMLV videos at the native frame rate of 200 fps, using the official pretrained BoostingVRME models. Every source frame is preserved throughout magnification, generation, demagnification, and evaluation.

We report spotting $F1_{spot}$, recognition $F1_{rec}$, and Spot-then-Recognize Score (STRS), which jointly measures detection and classification performance, reflecting whether micro-expressions are both localized and correctly recognized:
$$
\mathrm{STRS}
=
\mathrm{F1}_{\mathrm{spot}}
\cdot
\mathrm{F1}_{\mathrm{rec}}.
$$

Table~\ref{tab:sammlv_spotting} compares full-video spotting performance across three frozen facial animation backbones, with and without the MEGMAG magnify--generate--demagnify pipeline.

MEGMAG consistently outperforms all three backbones across every reported metric. Improvements in spotting F1 and STRS indicate that magnifying micro-expression dynamics before animation and attenuating them afterward makes generated expressions more detectable in their original temporal context. These results complement the recognition findings, as successful classification of segmented clips does not necessarily translate to reliable detection in long, untrimmed videos.

\textbf{Comparison with State-of-the-Art Micro-Expression Generation Methods.}
We compare MEGMAG with two methods specifically designed for micro-expression generation: FPB-FOMM~\cite{zhang2021fpbfomm} and TPS-FaceParsing~\cite{Yu_2022_ACMMM}. Unlike these dedicated approaches, MEGMAG retains a frozen general-purpose facial animation backbone and adapts only the Magnification and DeMagnification Modules.

All methods are evaluated on the same CAS(ME)$^3$ test split using the same three-model BoostingVRME logit ensemble. Since neither baseline provides pretrained weights or a documented training configuration for SAMMLV, we restrict this comparison to CAS(ME)$^3$.

Table~\ref{tab:comparison_prior_micro_generation_compact} summarizes the results. MEGMAG outperforms both dedicated generation methods across all three recognition metrics (UAR, UF1, and Accuracy), consistently achieving stronger performance with each of the three frozen animation backbones.

\textbf{Ablation Study}
Table~\ref{tab:emoportraits_ablation} presents an ablation study of MEGMAG using the frozen EmoPortraits backbone, evaluated on CAS(ME)$^3$ and SAMMLV through micro-expression recognition.
The first row shows the original EmoPortraits baseline. The second applies generic FlowMag motion magnification~\cite{Pan2023SelfSupervised} before EmoPortraits. The third replaces FlowMag with our learned Magnification Module. The final row reports the complete MEGMAG pipeline, combining our Magnification Module, the frozen EmoPortraits backbone, and the DeMagnification Module.

MEGMAG achieves the best or second-best performance across all metrics on both datasets. On CAS(ME)$^3$, the complete framework outperforms all other configurations on all three metrics. Compared with generic FlowMag, our learned Magnification Module improves performance across both datasets and all metrics, except for Accuracy on CAS(ME)$^3$.

Adding the DeMagnification Module yields the strongest overall results, outperforming all other configurations on every metric except UAR on SAMMLV, where magnification alone performs best, and Accuracy on SAMMLV, where the complete framework ranks second.

\begin{table}[t]
\centering
{\scriptsize \setlength{\tabcolsep}{2pt} \renewcommand{\arraystretch}{1.05}
\begin{tabular*}{\columnwidth}{@{\extracolsep{\fill}}lccc@{}}
\toprule
\textbf{Method}
& \textbf{UAR} $\uparrow$
& \textbf{UF1} $\uparrow$
& \textbf{Acc.} $\uparrow$ \\
\midrule

FPB-FOMM~\cite{zhang2021fpbfomm}
& 0.3630
& 0.3410
& 0.4221 \\

TPS-FaceParsing~\cite{Yu_2022_ACMMM}
& 0.2937
& 0.2768
& 0.3769 \\

\midrule

MEGMAG+FOMM
& 0.3856
& \underline{0.3737}
& \underline{0.4874} \\

MEGMAG+MetaPortrait
& \underline{0.3878}
& 0.3292
& 0.4623 \\

MEGMAG+EmoPortraits
& \textbf{0.4344}
& \textbf{0.4125}
& \textbf{0.5276} \\

\bottomrule
\end{tabular*}
}
\caption{Micro-expression generation comparison of MEGMAG against FPB-FOMM~\cite{zhang2021fpbfomm} and TPS-FaceParsing~\cite{Yu_2022_ACMMM} on the CAS(ME)\textsuperscript{3}~\cite{Li2023CASME3} test split, using BoostingVRME~\cite{Guo2025BoostingVRME} to evaluate recognition performance.}
\label{tab:comparison_prior_micro_generation_compact}
\end{table}

\section{Conclusions and Limitations}
\label{sec:conclusions}
We introduced MEGMAG, a magnify--generate--demagnify framework that enables pretrained facial animation backbones to generate micro-expressions without backbone retraining or architectural modifications. The Magnification Module amplifies subtle driver motion before expression transfer, while the DeMagnification Module attenuates the generated motion to restore realistic micro-expression intensity. Only these two modules are fine-tuned to the target micro-expression domain, while the animation backbone remains frozen.
Experiments demonstrate consistent improvements in micro-expression recognition and full-video spotting compared with unmodified backbones and existing micro-expression generation methods. Ablation studies further validate the contributions of the learned Magnification Module and the complete magnify--generate--demagnify pipeline.

A limitation of MEGMAG is the need to select magnification and attenuation factors for each backbone. Future work will explore automatic factor selection to reduce validation requirements and improve generalization across datasets and animation models.

\begin{table}[t]
\centering
{\scriptsize
\setlength{\tabcolsep}{2pt}
\renewcommand{\arraystretch}{1.05}
\begin{tabularx}{\columnwidth}{@{}>{\raggedright\arraybackslash}Xccc@{\hspace{5pt}}ccc@{}}
\toprule
\multicolumn{1}{c}{Dataset:}
& \multicolumn{3}{c}{\textbf{CAS(ME)\textsuperscript{3}}}
& \multicolumn{3}{c}{\textbf{SAMMLV}} \\
\cmidrule(lr){2-4}\cmidrule(lr){5-7}
\multicolumn{1}{c}{\textbf{Method}}
& \textbf{UAR$\uparrow$}
& \textbf{UF1$\uparrow$}
& \textbf{Acc.$\uparrow$}
& \textbf{UAR$\uparrow$}
& \textbf{UF1$\uparrow$}
& \textbf{Acc.$\uparrow$} \\
\midrule

EmoPortraits
& 0.3962 & 0.3645 & 0.4523
& 0.2727 & \underline{0.2609} & \textbf{0.6429} \\

FlowMag+EmoPortraits~\cite{Pan2023SelfSupervised}
& 0.4050
& 0.3679
& \underline{0.4807}
& 0.2941
& 0.2439
& 0.5769 \\

MagnificationModule+EmoPortraits
& \underline{0.4148}
& \underline{0.3701}
& 0.4724
& \textbf{0.3111}
& 0.2456
& 0.5833 \\

\midrule

MEGMAG+EmoPortraits
& \textbf{0.4344}
& \textbf{0.4125}
& \textbf{0.5276}
& \underline{0.3077}
& \textbf{0.2857}
& \underline{0.6316} \\

\bottomrule
\end{tabularx}
}

\caption{Micro-expression ablation study evaluating magnification and demagnification on frozen EmoPortraits backbone. We compare the original backbone, FlowMag~\cite{Pan2023SelfSupervised}+EmoPortraits, our learned Magnification Module+EmoPortraits, and the complete MEGMAG pipeline. Recognition performance is evaluated on CAS(ME)\textsuperscript{3}~\cite{Li2023CASME3} and SAMMLV~\cite{yap2020sammlv}.}
\label{tab:emoportraits_ablation}
\end{table}

\bibliographystyle{ieee}
\bibliography{base}

\end{document}